\documentclass[preprint,12pt]{elsarticle}
\usepackage[utf8]{inputenc}
\usepackage[T1]{fontenc}

\usepackage{amssymb}

\journal{Computer Methods and Programs in Biomedicine}

\usepackage{amsmath,amsfonts}
\usepackage{algorithmic}
\usepackage{xcolor}

\usepackage{bbm}
\usepackage{svg}

\usepackage{pgfplots}
\pgfplotsset{compat=1.18}	
\usetikzlibrary{pgfplots.statistics} 
\usetikzlibrary{matrix}
\usepgfplotslibrary{groupplots}
\usepackage{siunitx}

\usepackage{tikz}
\usetikzlibrary{positioning, calc, chains}
\usetikzlibrary{decorations, decorations.pathreplacing, decorations.pathmorphing, calligraphy}
\usetikzlibrary{arrows.meta}
\usetikzlibrary{
    shapes,
    shapes.geometric,
    shapes.symbols,
    shapes.arrows,
    shapes.multipart,
    shapes.callouts,
    shapes.misc}
     
\tikzset{>=Stealth[round]}
\tikzset{
    ncbar angle/.initial=90,
    ncbar/.style={
        to path=(\tikztostart)
        -- ($(\tikztostart)!#1!\pgfkeysvalueof{/tikz/ncbar angle}:(\tikztotarget)$)
        -- ($(\tikztotarget)!($(\tikztostart)!#1!\pgfkeysvalueof{/tikz/ncbar angle}:(\tikztotarget)$)!\pgfkeysvalueof{/tikz/ncbar angle}:(\tikztostart)$)
        -- (\tikztotarget)
    },
    ncbar/.default=0.5cm,
}

\tikzset{square left brace/.style={ncbar=0.2cm}}
\tikzset{square right brace/.style={ncbar=-0.2cm}}

\usepackage{hyperref}
\usepackage{array}
\usepackage{multirow}
\usepackage{color, colortbl}
\usepackage{comment}
\usepackage{booktabs}

\definecolor{Vgreen}{RGB}{141,211,199}
\definecolor{Vviolet}{RGB}{190,186,218}
\definecolor{Vred}{RGB}{251,128,114}
\definecolor{Vblue}{RGB}{128,177,211}
\definecolor{Vorange}{RGB}{253,180,98}
\definecolor{Vlime}{RGB}{179,222,105}
\definecolor{Vpink}{RGB}{252,205,229}
\definecolor{Vpurple}{RGB}{188,128,189}

\hypersetup{
  colorlinks=false,
  linkbordercolor=white,
  urlbordercolor=white,
  pdfborder={0 0 0}
}

\graphicspath{{./img/}}

\begin{document}

\begin{frontmatter}



\title{A Deep Learning Approach for Overall Survival Prediction in Lung Cancer with Missing Values}


\author[aff1]{Camillo Maria Caruso\texorpdfstring{\fnref{contrib}}{}}
\ead{camillomaria.caruso@unicampus.it}
\author[aff1]{Valerio Guarrasi\texorpdfstring{\fnref{contrib}}{}}
\ead{valerio.guarrasi@unicampus.it}
\author[aff2]{Sara Ramella}
\ead{s.ramella@policlinicocampus.it}
\author[aff1,aff3]{Paolo Soda\texorpdfstring{\corref{cor}}{}}
\ead{p.soda@unicampus.it, paolo.soda@umu.se}

\cortext[cor]{Correspondence: paolo.soda@umu.se, p.soda@unicampus.it; Tel.: +39~06~22541~9622}
\fntext[contrib]{These authors contributed equally to this work.}

\affiliation[aff1]{organization={Research Unit of Computer Systems and Bioinformatics, Department of Engineering, Università Campus Bio-Medico di Roma},
            city={Rome},
            state={Italy},
            country={Europe}}

\affiliation[aff2]{organization={Operative Research Unit of Radiation Oncology, Fondazione Policlinico Universitario Campus Bio-Medico},
            city={Rome},
            state={Italy},
            country={Europe}}

\affiliation[aff3]{organization={Department of Diagnostics and Intervention, Radiation Physics, Biomedical Engineering, Umeå University},
            city={Umeå},
            state={Sweden},
            country={Europe}}

\begin{abstract}

\textit{Background and Objective:} 
In the field of lung cancer research, particularly in the analysis of overall survival (OS), artificial intelligence (AI) serves crucial roles with specific aims.
Given the prevalent issue of missing data in the medical domain, our primary objective is to develop an AI model capable of dynamically handling this missing data.
Additionally, we aim to leverage all accessible data, effectively analyzing both uncensored patients who have experienced the event of interest and censored patients who have not, by embedding a specialized technique within our AI model, not commonly utilized in other AI tasks.
Through the realization of these objectives, our model aims to provide precise OS predictions for non-small cell lung cancer (NSCLC) patients, thus overcoming these significant challenges.

\textit{Methods:} We present a novel approach to survival analysis with missing values in the context of NSCLC, which exploits the strengths of the transformer architecture to account only for available features without requiring any imputation strategy. 
More specifically, this model tailors the transformer architecture to tabular data by adapting its feature embedding and masked self-attention to mask missing data and fully exploit the available ones. 
By making use of ad-hoc designed losses for OS, it is able to account for both censored and uncensored patients, as well as changes in risks over time.

\textit{Results:} We compared our method with state-of-the-art models for survival analysis coupled with different imputation strategies. 
We evaluated the results obtained over a period of 6 years using different time granularities obtaining a \textit{Ct-index}, a time-dependent variant of the \textit{C-index}, of 71.97, 77.58 and 80.72 for time units of 1 month, 1 year and 2 years, respectively, outperforming all state-of-the-art methods regardless of the imputation method used. 

\textit{Conclusions:} The results show that our model not only outperforms the state-of-the-art’s performance but also simplifies the analysis in the presence of missing data, by effectively eliminating the need to identify the most appropriate imputation strategy for predicting OS in NSCLC patients.

\end{abstract}




\end{frontmatter}

\section{Introduction}\label{sec:introduction}

In recent years, Artificial Intelligence (AI) has rapidly become a part of our daily lives, including healthcare~\cite{bib:review_health}. 
There have been notable advances in applying quantitative methods in clinical practice, which have paved the way for precision medicine. 
Cancer research is one of the most promising areas where AI can be applied~\cite{bib:review_cancer}. 
According to the World Health Organization, cancer is responsible for nearly 10 million yearly deaths globally, with lung cancer accounting for 18\% of these~\cite{bib:WHO} and non-small cell lung cancer (NSCLC) is the most frequent type, being approximately 82\% of all cases~\cite{bib:cancer}.

In cancer research, a key factor is the overall survival (OS) of the patient, which refers to the time from the initial cancer diagnosis to the time of death. 
Identifying subgroups of patients with a higher or lower chance of survival is critical in developing effective strategies to improve OS rates. 
For example, in the case of NSCLC, the 5-year survival rate is only 26\%, and this rate drops to a mere 7\% when cancer returns locally or spreads to distant organs~\cite{bib:cancer}. 
Therefore, although some AI-based methods have been proposed in healthcare~\cite{bib:reviewOS}, accurately predicting OS remains a major challenge. 

In medicine, it can be quite challenging to obtain a dataset, i.e., tabular dataset, with no missing features: how to handle this situation effectively is a key question in AI research. 
In fact, traditional methods require complete data samples, so standard practices involve either excluding samples with missing values or utilizing imputation strategies. 
However, it is important to note that these methods can compromise the findings, introducing bias and reducing statistical power. 
Therefore, research should go towards the development of models which can cope with missing features without any imputation. 

To address these issues, we introduce a novel approach that tackles the problem of handling missing features in tabular data in the study of OS analysis for patients affected by NSCLC.
Indeed, tabular data are the most simple and diffused data type in AI applications~\cite{bib:tabular}.
Our method stems from the transformer architecture~\cite{bib:transformer} and leverages the idea of the mask inside the self-attention module to learn from incomplete input data.
By masking missing features, it learns only from available data, avoiding any type of imputation of missing features, with the goal of improving the performance of survival prediction of patients affected by NSCLC.

To validate the proposed model, we compared its predictive performance with feature imputation and OS prediction state-of-the-art approaches. 
Indeed, even though some methods are capable of handling missing features in the test phase, to the best of our knowledge, none has yet addressed the issue of how to deal with them during training without the need for any imputation method in the context of OS prediction~\cite{bib:reviewOS}.

The main contributions of this work are the following:
\begin{itemize}
    \item We propose a specialized transformer-based decision support system on tabular data, for predicting lung cancer OS using clinical data.
    \item Our approach efficiently handles missing data, eliminating the need for imputation strategies.
    \item Our approach uses ad-hoc designed loss functions allowing the use of both censored and uncensored patients, without the need to exclude any patient.
    \item We assessed the performance of the proposed model at various time granularities, confirming its robustness across different time windows. 
    The results obtained on the OS prediction in NSCLC patients outperform OS state-of-the-art models, regardless of the imputation method used.
\end{itemize}

The manuscript is organized as follows: section~\ref{sec:background} presents the state-of-the-art of survival analysis and data imputation techniques; section \ref{sec:materials} reports the details about the data employed in the analyses; section~\ref{sec:methods} introduces the proposed model and explains the metrics employed; section~\ref{sec:results} discusses the experimental results; section~\ref{sec:conclusion} provides concluding remarks.

\section{Background}\label{sec:background}

Survival analysis, also known as time-to-event analysis, plays a crucial role in various fields, especially medicine~\cite{bib:surv1}.
It aims to understand the relationship between covariates, such as patient features, and the distribution of survival times. 
In the lung cancer setting, this type of analysis helps to identify risk factors that affect survival and to compare risks among different subjects, with the aim of tailoring the “right therapy for the right patient".

One important aspect to consider in this particular field is censored data, specifically right-censoring. 
This occurs when a patient withdraws from the study, is lost to follow-up, or is still alive without experiencing the event of interest at the last follow-up. 
In such cases, a special analysis is required since it is not known when and if the event occurred, and thus these samples cannot be considered together with uncensored data, i.e., patients for which it is known when they have experienced the event. 
However, most of the studies that predict OS in NSCLC have approached the survival problem as a classification task, dividing patients into high and low risks based on a survival time threshold~\cite{bib:thresh1, bib:thresh2, bib:thresh3, bib:thresh4, bib:thresh5, bib:CMPB2}.
This approach does not fully exploit the information of the censored patients, since it excludes from the analysis those patients with a survival time shorter than the threshold.
In contrast, other studies predict the risks faced by patients, which can be used to evaluate the correct ordering of the patients through the \textit{C-index} metric~\cite{bib:order1, bib:order2, bib:order3, bib:order4, bib:CMPB1, bib:CMPB3}. 
The downside of these approaches is that they do not make full use of the available temporal information, not taking into account possible changes in risk over time.
On the contrary, another possible approach is to use several output nodes for each time interval, which represent the risk the patient experienced at the specific time interval.
Indeed, in \cite{bib:deephit} the authors present ad-hoc designed loss functions to exploit information from both uncensored and censored patients, enabling learning from the latter that the event of interest did not occur up until the last follow-up.
This was achieved by defining the survival problem as identifying the first time an underlying stochastic process hits a specific boundary, also known as the first hitting time. 
Although this approach still does not enable predictions to be made about censored samples, it does permit learning from them. 

To tackle the survival task, a few models in the literature have been proposed.
The most standard used approach is the Kaplan-Meier estimator, which can take censored samples into account, but does not incorporate patient covariates, making it useful to estimate the survival rate at the population level but not at the patient level. 
Instead, the most commonly used method is the Cox proportional hazard~(CPH), which incorporates the features of the patient, but assumes that the hazard rate, i.e., the probability of experiencing the event within a short time interval, is constant and that the log of the hazard rate is a linear function of the covariates.
These two assumptions are known as the proportional hazard assumption, and a few methods have been proposed to address the limitations it introduces. 
The Survival Tree~(ST) and the Random Survival Forest~(RSF) are extensions of the CART and Random Forest algorithms that deal with censored survival data. 
ST accomplished this by a recursive partitioning procedure based on maximizing the dissimilarity in the survival distributions of patients between different regions of the covariate space, thus taking into account not only the homogeneity of predictor variables but also the survival probability and time of the events. 
RSF estimates the cumulative hazard function for each case's terminal nodes by leveraging out-of-bag data and then averages the cumulative hazard functions of all trees in the forest. 
The cumulative hazard function is a measure that describes the cumulative probability of experiencing a particular event, such as death or failure, up to a specified time point.
Instead, DeepHit~(DH)~\cite{bib:deephit} offers a new approach that predicts the probability of the first hitting time of an event using a deep neural network, employing a loss function that exploits survival times and relative risks to predict the cumulative incidence function, a measure useful in understanding the risk of developing a particular health outcome, as it takes into account both the risk of developing the outcome and the time period over which the risk is evaluated.

When working with medical data, an important aspect to consider is the presence of incomplete records. 
This phenomenon may be due to different reasons and can hinder the employment of AI models.
Although some strategies have been proposed to address this issue, only a few methods are commonly used in a healthcare setting~\cite{bib:reviewOS}: complete case analysis, overall mean imputation, k-nearest neighbors (kNN) imputation, multiple imputations by chained equations (MICE) and MissForest.
The first two methods for dealing with missing data, complete case analysis, and overall mean imputation, involve either discarding incomplete samples or imputing the missing values with the mean value of the corresponding feature computed from the available data, which are straightforward but have some drawbacks.
Specifically, the former may discard too many samples, which can prevent the learning of deep learning techniques, whereas the latter has a high likelihood of introducing bias into the model's final outcome.
As a result, kNN, MICE and MissForest have been proposed as alternative solutions. 
The first method identifies similar samples to the one being analyzed by calculating the distance based on other non-missing features in order to impute a more appropriate value.
The second one imputes missing values by modeling each feature with missing values as a function of other features in a round-robin fashion. 
The third one uses a Random Forest model to predict missing data, which starts by imputing missing values using a simple strategy, e.g., mean and mode imputation, and then iteratively updates the imputations using the model predictions, refining these estimates until the changes between iterations are minimal.

In this context, we present our approach, which combines the strengths of DH and transformer architecture~\cite{bib:transformer}. 
We use the transformer architecture to exploit the power of the mask in the self-attention module, which allows us to avoid any imputation of missing values in the survival analysis task.

\section{Materials}\label{sec:materials}

To validate our clinical decision support system, presented in the next section, we used clinical data from the CLARO dataset~\cite{bib:CLARO}.
This consists of 297 patients affected by NSCLC, who underwent concurrent chemoradiation for locally advanced NSCLC and systemic treatment for metastatic disease.  
The OS of the entire population, which included 184 censored and 113 uncensored patients, has a mean of 20.74$\pm$42.45 months (95\%~CI).
The population was enrolled in the study under two separate Ethical Committee approvals, including a retrospective phase that was approved on October 30th, 2012, and registered on ClinicalTrials.gov on July 12th, 2018 with the identifier NCT03583723. 
The prospective phase was approved with the identifier 16/19 OSS. 
The Institutional Review Board approved this analysis, and all patients provided written informed consent.

The clinical data that we have collected contains 8 clinical descriptors, as outlined in \tablename~\ref{tab:features}, which reports the distribution and the amount of missing values for each feature.
As mentioned above, our features are composed of personal information, i.e., \textit{Age} and \textit{Sex} (assigned at birth), and details about tumor histopathology, i.e., clinical target volume (\textit{CTV}), \textit{Overall Stage},  tumor ~(\textit{T}), nodule~(\textit{N}), metastasis~(\textit{M}) stages and \textit{Histology}.
To determine the stage of the tumor, we had two radiation oncologists with extensive experience independently review CT scans and assign staging scores for the tumor: \textit{Overall Stage}, \textit{T}, \textit{N}, \textit{M}. 
If there was any disagreement between the two experts, they would review the patient's CT images together until they reached a consensus.
It is worth noting that some patients in our study did not undergo a histopathological examination, this is why the “unknown" class is included as one of the categories for the \textit{Histology} feature.

To prepare the data for analysis, we applied one-hot encoding for categorical features and z-score normalization for continuous features\footnotemark.
For our approach, we generated empty vectors for the one-hot encoding of the missing features.
\footnotetext{This operation is performed on the test set based on the parameters computed using the training set.}

\begin{table}[!ht]
\centering
\scalebox{.75}{
\setlength{\extrarowheight}{2pt}
\begin{tabular}{cccc}
\hline
\textbf{Feature} & \textbf{Missing Data} &           \textbf{Categories} & \textbf{Distribution} \\
\hline
\textit{Age*}       &     0 (0.0\%) &      $<68~years$  & 138~(46.46\%) \\
                   &               &   $\geq 68~years$  & 159~(53.54\%) \\[.2cm]
\textit{Sex}    &   0 (0.0\%) &                    F  & ~97~(32.66\%) \\
                   &               &                  M  & 200~(67.34\%) \\[.2cm]
\textit{CTV*}       & 112 (37.71\%) &     $<146.51~cm^3$ & 115~(38.72\%) \\
                   &               & $\geq 146.51~cm^3$  & ~70~(23.57\%) \\[.2cm]
\textit{Overall Stage}     &     0 (0.0\%) &                 II  & ~~8~(2.69\%)~ \\
                   &               &                III  & 188~(63.31\%) \\
                   &               &                 IV  & ~96~(32.32\%) \\
                   &               &         Recurrence  & ~~5~(1.68\%)~ \\[.2cm]
\textit{T}   & 116 (39.06\%) &                  1  & ~11~(3.70\%)~ \\
                   &               &                  2  & ~40~(13.46\%) \\
                   &               &                  3  & ~74~(24.92\%) \\
                   &               &                  4  & ~56~(18.86\%) \\[.2cm]
\textit{N}   & 104 (35.02\%) &                  0  & ~17~(5.72\%)~ \\
                   &               &                  1  & ~39~(13.13\%) \\
                   &               &                  2  & 112~(37.71\%) \\
                   &               &                  3  & ~18~(6.06\%)~ \\
                   &               &         Recurrence  & ~~7~(2.36\%)~ \\[.2cm]
\textit{M}   &  ~94 (31.65\%) &                 0  & 201~(67.68\%) \\
                   &               &                  1  &  ~~2~(0.67\%)~ \\[.2cm]
\textit{Histology} &    ~3 (1.01\%) &     Adenocarcinoma & 153~(51.52\%) \\
                   &               &           Squamous  & ~73~(24.58\%) \\
                   &               &              Other  & ~20~(6.73\%)~ \\
                   &               &            Unknown  & ~48~(16.16\%) \\[.2cm]
\hline
\end{tabular}}
\caption{Patients' characteristics. For each feature is reported the number and percentage of missing values, along with the distribution (number and percentage) of the possible categorical values. 
Note that the variables marked with *, are continuous, but, for the sake of clarity, we have presented their distribution using their mean values as thresholds. 
It is important to note that the model used the original continuous values of these variables.}\label{tab:features}
\end{table}

\section{Methods}\label{sec:methods}

In this section, we first describe the architecture of the proposed approach with an example of its main blocks, next we illustrate the loss function used during training and the evaluation metric employed, and finally we report the experimental setup used to perform the experiments. 

\subsection{Model}\label{sec:model}

Given that AI methods for OS prediction require complete data~\cite{bib:reviewOS}, and that in healthcare it can be quite challenging to obtain a complete data set without missing values, we need to effectively handle the data without the necessity of imputing or removing any information.
Taking inspiration from the cutting-edge transformer architecture~\cite{bib:transformer}, we present here a novel model that takes into account only the available features. 
Our approach involves adapting the transformer's encoder architecture to tabular data, via a novel positional encoding for tabular features, and utilizing padding to mask any missing features within the attention module, enabling the model to ignore them effectively. 

\begin{figure}[!ht]
\centering
\includegraphics[width=.85\columnwidth, keepaspectratio]{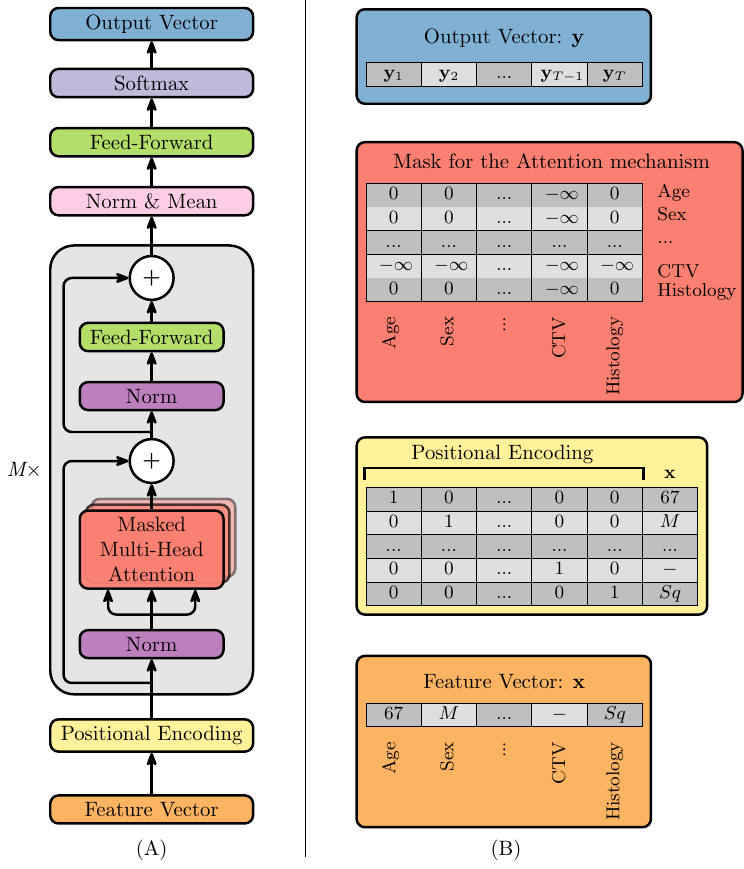}
\caption{Schematic representation of the proposed model: (A) Architecture of the proposed approach and (B) example of positional encoding, mask and output vector, where the $-$ symbol represents a missing feature. 
Note that for simplicity of representation, just a few features are reported and none of the preprocessing procedures are applied. }
\label{fig:model}
\end{figure}

The schematic representation of the proposed model is shown in \figurename~\ref{fig:model}: left panel, denoted by the letter A, represents the overall architecture from the input (the orange block shown at the bottom of the figure) to the output (the blue block at the top of the figure), whilst right panel, marked as B, offers an example of four blocks of panel A, which can be identified by the name as well as by the color. 

The input to the model, shown in the first block represented in orange in \figurename~\ref{fig:model}, is a feature vector $\mathbf{x}$ of dimension $d$ composed of the preprocessed patient information.
The second block, represented in yellow, is the positional encoding, which is used for identifying the feature itself without explicitly encoding the feature order.
To achieve this, we explored the use of a one-hot encoding vector representing the position of each feature. 
In this respect panel B of \figurename~\ref{fig:model} shows an example in the case of the CLARO dataset: here the initial feature vector is turned into a $d\times (d+1)$ matrix, where the columns from the first to the second-last represent the positional encoding, whilst the last column reports the feature vector.

The third block is the transformer encoder (light gray block), which consists of a stack of $M$ identical layers. 
Each layer has two sub-layers: the first is a multi-head self-attention mechanism (red block), and the second is a position-wise fully connected feed-forward network (green block). 
As in the original architecture~\cite{bib:transformer}, we employ a layer normalization (purple block) and a residual connection (denoted with $\oplus$) around each of the two sub-layers.
The normalization step precedes each of these blocks to reduce the risk of the vanishing or exploding gradients problem, which can occur during the training process.

To handle the presence of missing features, we drew inspiration from the padding mask technique used in natural language processing~\cite{bib:transformer}. 
This technique extends the capabilities of the attention mechanism, preventing leftward information flow, which exploits the use of the mask within the scaled dot-product attention mechanism to mask out, setting to~$-\infty$, any values that would result in illegal connections.
In this regard, the red block of panel B of \figurename~\ref{fig:model} shows an example of the mask employed in the attention mechanism.
Using such a mask, we were able to effectively ignore any missing features, without requiring any imputation strategy.

Next, we applied a normalization and averaging step (pink block) in order to get a latent representation vector, respective to the non-missing features, to be fed to the final classifier, composed of a feed-forward module (green block) and softmax function (violet block).
The feed-forward module that succeeds the encoder maps the encoder's embedding to the output vector's dimension \textit{T}.
Hence, each element $\mathbf{y}_t$ of the output vector $\mathbf{y}$ (blue block) represents the probability that the event occurs at the time point $t$.

\subsection{Training and Testing}\label{sec:train_test}

When training the model, the goal is to correctly predict the probability that the event of interest, formally denoted as $k=1$ and corresponding to the patient's death, occurs at time $t$, given the patient's feature vector $\mathbf{x}$ and under the constraint that $t$ is smaller or equal to $s$, the true time when the event $k=1$ occurs.
Straightforwardly, we identify with $k=0$ the event of censoring for patients who did not experience the event.
The cumulative incidence function $F(t|\mathbf{x})$ is therefore defined as:

$$
    F(t|\mathbf{x}) = P(s \leq t, k = 1 | \mathbf{x}) = \sum_{s = 0}^{t} P( s , k = 1 | \mathbf{x})
$$

However, since the true cumulative incidence function $F(t|\mathbf{x})$ is not known, it can be estimated with $\hat{F}(s|\mathbf{x})$, which is the cumulative sum of the outputs of the model, defined as:

$$
    \hat{F}(s|\mathbf{x}) = \sum_{t=0}^{s} \mathbf{y}_t.
$$

Therefore we can now compute this measure and thus compare the risks faced by different patients at a specific time. 

We train the proposed model using the loss function $L$ presented in \cite{bib:deephit}, specifically designed to handle censored patients.
It is composed of two terms, $L_1$ and $L_2$, so that 

$$
L = L_1 + L_2
$$

where the first term captures the death and censoring times of patients, evaluating respectively the output $\mathbf{y}$ and the estimated cumulative incidence function $\hat{F}$ at patient-specific event times, whereas the second term measures the correct ordering of the patients based on their relative risk.
Indeed, $L_1$ is the log-likelihood of the distribution of the first hitting time of an event, defined as:

\begin{align*}
    L_1 = &- \sum_{i=1}^N [\mathbbm{1}(k^{(i)} = 1) \cdot \log(\mathbf{y}_{s^{(i)} }^{(i)})~+ \\ &+ \mathbbm{1}(k^{(i)} = 0) \cdot \log(1-\hat{F}(s^{(i)}|\mathbf{x}^{(i)})) ]
\end{align*}

where $\mathbbm{1}({\cdot})$ is an indicator function, and the summation is performed on all $N$ samples, where the apex $(i)$ indicates that the information is related to the specific patient $i$.
Through two terms, it exploits the information of both uncensored and censored patients: the first term captures the first hitting time to maximize the patient's risk at the specific time of occurrence of the event; the second one maximizes the survival function, defined as $1 - F(t)$~\cite{bib:survival_function}, evaluated at the patient's last follow-up.

$L_2$ employs a ranking loss function based on the concept of concordance. 
Essentially, it indicates that a patient who has passed away at a certain time should be considered at a greater risk than a patient who is still alive at that time.
It is defined as: 

$$    
L_2 = \sum_{i=1}^N \sum_{j \neq i}^N A_{i,j} \cdot \exp\left(-\frac{\hat{F}(s^{(i)}|\mathbf{x}^{(i)}) - \hat{F}(s^{(i)}|\mathbf{x}^{(j)})}{0.1}\right)
$$

where $A_{i,j} = \mathbbm{1}\left(k^{(i)}=1, s^{(i)} < s^{(j)}\right)$. 
Using these terms the loss function penalizes the incorrect ordering of pairs.
Indeed, on the one hand, $A_{i,j}$ is an indicator function that identifies pairs of patients whose comparison is meaningful, known as acceptable pairs.
These pairs consist of patients in whom the first patient experienced the event at a specific time point, while the second patient has a longer survival time.
On the other hand, the exponential function compares the risks faced by the two patients, $i$ and $j$, at the time the first one experienced the event. 
In particular, it assumes small values in the case of a correct ordering, i.e. if the first patient has a higher risk than the second, whilst it assumes large values in the case of a wrong ordering.

\subsection{Evaluation Metric}\label{sec:metric}

Once a survival analysis model is trained, we need to evaluate its outputs taking time into account, as does the time-dependent concordance index (\textit{Ct-index}) does.
It is an evolution of the commonly used \textit{C-index}; this latter is based on the assumption that patients who lived longer should be assigned lower risks than those who lived a shorter period.
However, the \textit{C-index} does not account for any potential changes in risk over time, whereas the \textit{Ct-index} is calculated by comparing pairs of patients, where one patient has experienced the event at a specific time while the other patient has neither experienced the event nor been censored by that time.
Formally, \textit{Ct-index} is defined as:

\begin{align*}
    \textit{Ct-index} &= P\left(\hat{F}(s^{(i)}|\mathbf{x}^{(i)}) > \hat{F}(s^{(i)}|\mathbf{x}^{(j)})| s^{(i)} < s^{(j)}\right) \\
    &\approx \frac{\sum_{i=1}^N\sum_{j \neq i}^N A_{i,j} \cdot \mathbbm{1}\left(\hat{F}(s^{(i)}|\mathbf{x}^{(i)}) > \hat{F}(s^{(i)}|\mathbf{x}^{(j)})\right)}{\sum_{i=1}^N\sum_{j \neq i}^N A_{i,j}}
\end{align*}

This index is not based on a fixed time, unlike the \textit{C-index}, which considers only the ordering of subjects at a fixed time point. 
Instead, the \textit{Ct-index} takes into account the timing of events for each subject over time. 
In fact, using $A_{i,j}$ and the indicator function that verifies whether, at the time the first patient experienced the event, he/she presents a higher risk than the second one, this metric computes the fraction of correctly ordered patients out of the total number of acceptable pairs.

\subsection{Experimental setup}\label{sec:experimental_setup}

Our architecture adopts 12 consecutive encoder layers (\textit{M}), with each layer utilizing 17 attention heads and a feed-forward module composed of one hidden layer of 3072 neurons. 
We do not further investigate any other hyperparameters configuration, since their tuning is out of the scope of this manuscript.
Nevertheless, the “No Free Lunch" Theorem for optimization states that there is no universal set of hyperparameters that will optimize the performance of a model across all possible datasets~\cite{bib:NFL}.

To have a fair comparison between the models employed in the analysis, we applied 5-fold stratified cross-validation, maintaining the distribution of censored and uncensored patients among the different folds.
Thus, we divided the data into test set (20\%) and train set (80\%), part of which was used as validation set (20\%).

We compared our approach with each pair of state-of-the-art imputation strategies and models for OS analysis presented in section~\ref{sec:background}.
Regarding the imputation methods, we employed overall mean imputation strategy, kNN imputer, MICE and MissForest with their default parameters~\cite{bib:sklearn, bib:missforest}. 
More specifically, we opted for 5 neighbors and the euclidean distance for the kNN imputer, for a maximum number of iterations equal to 10 and a mean initial strategy for the MICE strategy, and for a maximum number of iterations equal to 10, 100 estimators and the squared error as a criterion for the growth of the MissForest.   
We opted not to use the complete case analysis since in our application the number of patients would be almost halved, passing from 297 to 158.
Focusing on the models, we tested the CPH, the ST, the RSF and the DH. 
For training the CPH, ST and RSF, we used their default parameters~\cite{bib:sksurv}. 
More specifically, we opted for the breslow method to handle tied event times and 100 iterations in the training of CPH, for the best splitter, a minimum of 6 samples to split a node and 3 samples to define leaf nodes in the growth of the ST, and 100 estimators, a minimum of 6 samples to split a node and 3 samples to define leaf nodes in the growth of the RSF. 
In training both our model and DH we opted for the Adam optimizer, a batch size of 32, an initial learning rate of $10^{-4}$ and a Xavier initialization.
The training was set to a maximum of 1500 epochs, in conjunction with an early stopping criterion and a learning rate scheduler both based on the validation loss with patience of 200 and 100 epochs, respectively.

To conduct the comparisons, we used different units of time, i.e., one month, one year, and two years, covering a period of six years.
Note that we defined this time limit in order to include at least 95\% of the patients' survival times without any modification, whereas we considered all the patients with a longer survival time as censored. 
Thus each element of the output vector indicates the level of risk that the patient faces within the corresponding time interval.
It is worth emphasizing that the output vector's size $N$, changes across the different experiments based on the specific unit of time employed: 72 elements for the 1-month, 6 for the 1-year and 3 for the 2-year time unit.


\section{Results and Discussions}\label{sec:results}

As reported in section~\ref{sec:experimental_setup}, we compared the performance of our approach with the state-of-the-art OS models, trained on the CLARO dataset~\cite{bib:CLARO}, using different imputation strategies.  
The results, in terms of \textit{Ct-index}, are presented in \tablename~\ref{tab:results}, where the first two columns report the combination of model and imputation method employed, whereas the rest of the columns represent the units of time applied.
The mean value and standard error reported are calculated on the different folds and the best mean performance for each unit of time is marked in bold.
As we can see, our approach always outperforms the benchmarks.

\begin{table}[ht]
\centering
\scalebox{.8}{
\setlength{\extrarowheight}{2pt}
\begin{tabular}{ccccc}
\toprule 
 \textbf{Model} & \textbf{Imputation} & \textbf{1-month} & \textbf{1-year} & \textbf{2-year} \\[.1cm]
\toprule 
CPH & Mean  & $61.10\pm3.05$*  & $60.92\pm3.59$*  & $47.64\pm8.72$ \\[.1cm] 
& kNN  & $60.72\pm3.90$  & $59.72\pm5.46$  & $48.51\pm9.14$ \\[.1cm]
& MICE  & $60.80\pm3.16$*  & $60.01\pm4.29$*  & $47.16\pm8.74$ \\[.1cm] 
& MissForest  & $61.34\pm3.05$*  & $58.95\pm4.11$*  & $44.99\pm9.44$ \\[.1cm] 
\midrule
ST & Mean  & $23.46\pm4.24$*  & $39.27\pm4.02$*  & $43.82\pm2.78$* \\[.1cm]
& kNN  & $27.05\pm3.14$*  & $22.51\pm3.20$*  & $35.28\pm4.29$* \\[.1cm]
& MICE  & $22.03\pm4.17$*  & $45.51\pm4.02$*  & $37.29\pm7.47$* \\[.1cm]
& MissForest  & $22.31\pm3.48$*  & $28.45\pm8.12$*  & $37.15\pm10.88$* \\[.1cm]
& $-$  & $26.49\pm2.05$*  & $37.61\pm7.32$*  & $42.71\pm6.19$* \\[.1cm]
\midrule
RSF & Mean  & $65.88\pm2.37$  & $72.16\pm3.02$  & $57.59\pm7.08$ \\[.1cm]
& kNN  & $61.92\pm3.27$*  & $68.24\pm1.34$*  & $54.66\pm8.10$ \\[.1cm]
& MICE  & $67.10\pm3.31$  & $71.53\pm4.06$  & $57.45\pm8.21$ \\[.1cm]
& MissForest  & $62.86\pm2.66$  & $67.69\pm5.99$  & $58.79\pm11.16$ \\[.1cm]
\midrule
DH & Mean  & $69.72\pm3.67$  & $75.19\pm5.38$  & $78.23\pm4.11$ \\[.1cm]
& kNN  & $68.24\pm3.94$  & $71.28\pm6.54$  & $77.94\pm3.66$ \\[.1cm] 
& MICE   & $71.04\pm2.92$  & $75.26\pm5.81$  & $78.39\pm3.54$ \\[.1cm]
& MissForest   & $68.39\pm3.41$  & $72.64\pm5.42$  & $74.64\pm5.34$ \\[.1cm]
\midrule
Ours & $-$ & $\mathbf{71.97\pm2.39}$ & $\mathbf{77.58\pm1.82}$ & $\mathbf{80.72\pm9.11}$ \\[.1cm]
\bottomrule 
\end{tabular}}
\caption{Performance of tested models in terms of \textit{Ct-index} (mean $\pm$ standard error). The asterisks (*) denote experiments with performances that are statistically different (\textit{p-value}$<0.05$) from our proposed model.}\label{tab:results}
\end{table}

\begin{figure}[!ht]
    \centering
    \resizebox{\linewidth}{!}{
    \begin{tikzpicture}
        \node[label=below: \tiny{(C)}] (C) at (0,0) {\includegraphics[height=5cm]{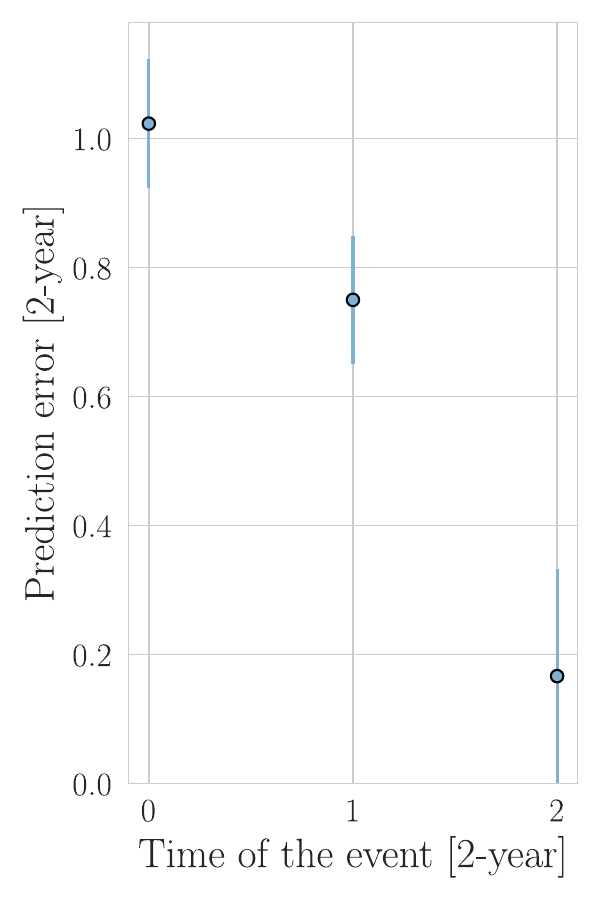}};
        \node[label=below: \tiny{(B)}] (B) [left=-.2cm of C] {\includegraphics[height=5cm]{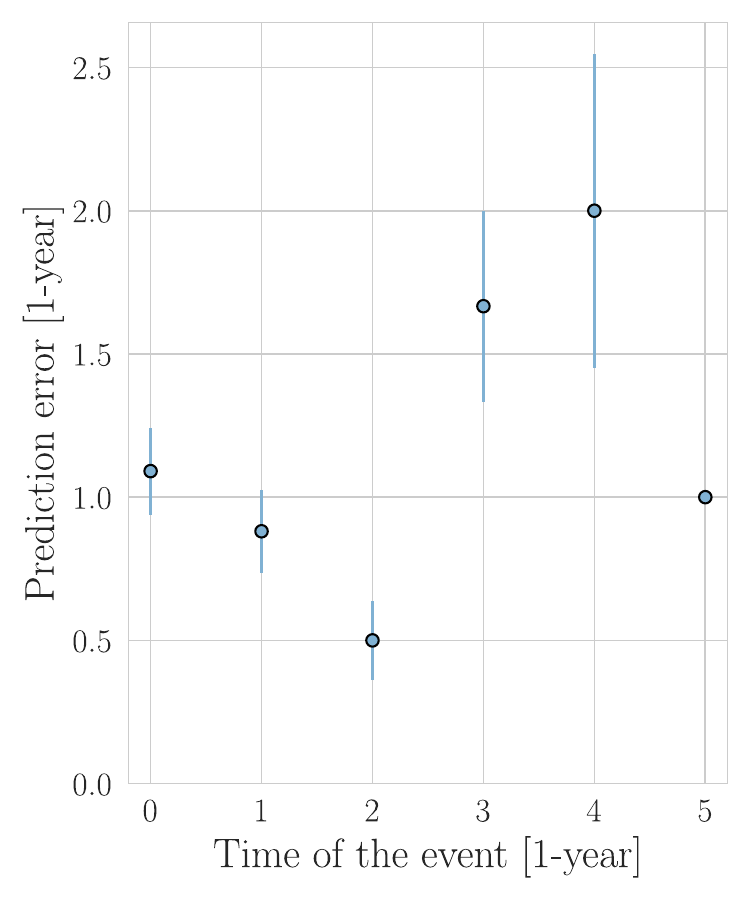}};
        \node[label=below: \tiny{(A)}] (A) [left=-.2cm of B] {\includegraphics[height=5cm]{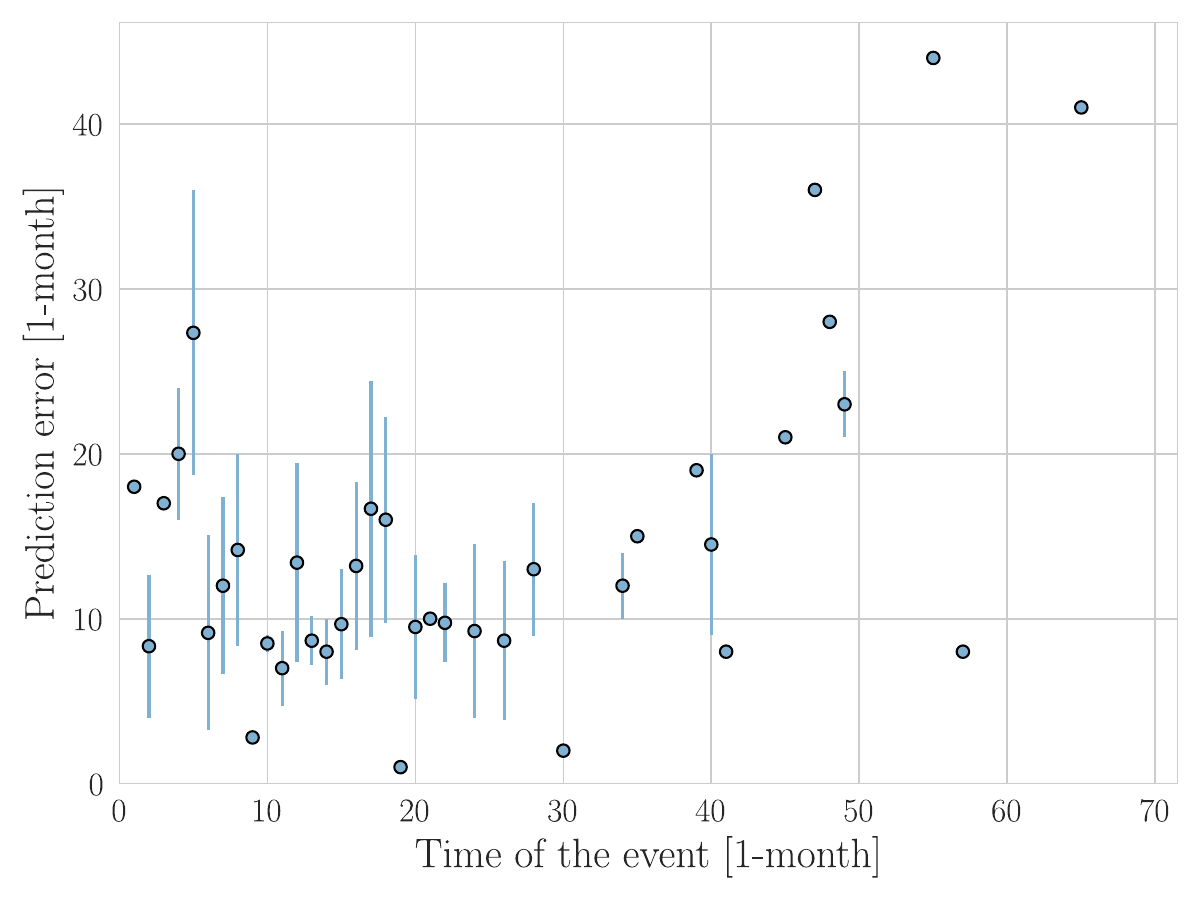}};
    \end{tikzpicture}}
    \caption{Prediction errors made for different patient groups by the proposed method, taking into account the actual times of occurrence of the event. The graph reports the mean and standard error of the prediction of uncensored patients, as for censored patients no valid prediction can be made since the event did not occur.}
    \label{fig:errors}
\end{figure}

These results, independently from the imputation strategy applied, confirm the considerations made in section~\ref{sec:background}. 
It appears that CPH performs the worst, likely due to the limitations of the proportional hazard assumption in this context.
Conversely, all other methods, which eliminate this constraint, perform better than CPH. 
Notably, DH and our approach outperform RSF, showcasing the potential of deep learning once again.
We deem that the improvement in the performance of our approach compared with DH is due to its ability to handle a high degree of missing data without the need for imputation, which has the potential to bias the final prediction.
Indeed, the missing data imputation itself poses some challenges in the selection of the most appropriate approach to the task at hand; furthermore, most of the existing imputation methods struggle to handle high levels of missing data, leading to a drop in performance.
On the contrary, our approach smoothly copes with this situation by learning only from the available features.

When we examine the \textit{Ct-index} of the different models at various units of time, we observe that those of DH and our model improve as unit time increases. 
This can be attributed to the reduction of complexity for the task of correctly ordering pairs, as the granularity of the problem reduces, suggesting that our method effectively interprets the available features to estimate the OS time.

Focusing on the differences between the various imputation strategies, \tablename~\ref{tab:results} highlights how difficult is to determine the most appropriate method, since the outcome depends both on the data and the models employed in the analysis. 
Therefore, our approach not only proves to be the best in terms of achieving the highest performance, but it also enables us to eliminate one of the variables from the problem by simply disregarding the missing features instead of searching for the most appropriate imputation strategy for the task at hand.

To further validate our results, we performed a paired t-test to evaluate the statistical difference between the performance obtained in different folds by our model compared with all competitors. 
In \tablename~\ref{tab:results}, denoted by the asterisks (*), we report the results with a \textit{p-value}$< 0.05$. 
These results show that our approach is significantly better than ST for all combinations with imputers and with all time granularities. 
Most notable, among the various combinations, is the case without imputers, showing that our model can handle missing values better. 
With regards to CPH, our model is statistically better than the configurations in combination with overall mean imputation, MICE and MissForest. 
Moreover, when the time granularity is increased to 2 years, the performance is no longer statistically distinguishable from those of our model, again indicating the reduction in complexity of the problem as granularity increases. 
In contrast, looking at the results obtained by RSF and DH, only the combination of RSF with kNN imputer for the time granularities of 1 month and 1 year are statistically worse than those of our model. 
Despite this, the results obtained from our model are still worthwhile since they are on average superior or comparable to the others, but having simplified the overall analysis, since no imputation of missing data is required.

Moreover, we performed an analysis to better understand on which patients the proposed model makes most errors. 
In \figurename~\ref{fig:errors}, in panels A, B, and C for the time granularities of 1 month, 1 year, and 2 years, respectively, we reported the mean errors and relative standard errors for uncensored patients grouped by unit of time in which the event occurred. 
From panel A we can observe that most uncensored patients, who present the event in the first few years after the diagnosis, have the lowest error, which is below 30 months, whereas, with regard to those patients who have survived longer, the errors are on average larger, even surpassing 40 months. 
When the granularity of time increases, on the other hand, the trend reverses, leading to a minimum error of $4.8$ months ($0.2$ of a 2-year period) in the case of patients who died in the third two-year period, as reported in panel C. 
This trend inversion, which probably also affects performance explaining its increase as time granularity increases, is perhaps due to the difference in the numbers of patients in the different groups of patients. 

\begin{figure}[!ht]
    \centering
    \resizebox{\linewidth}{!}{
    \begin{tikzpicture}
        \node[label=below: \tiny{(C)}] (C) at (0,0) {\includegraphics[height=5cm]{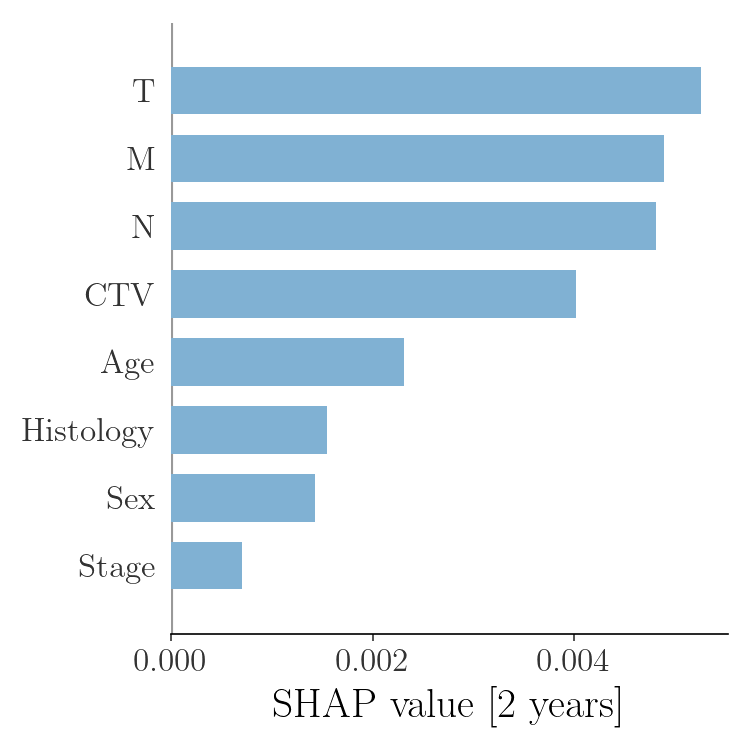}};
        \node[label=below: \tiny{(B)}] (B) [left=-.2cm of C] {\includegraphics[height=5cm]{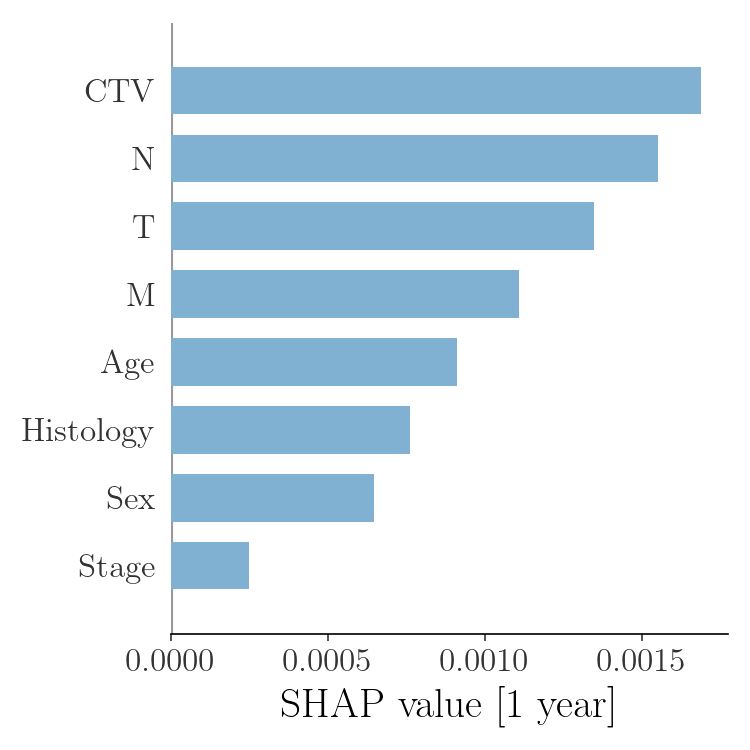}};
        \node[label=below: \tiny{(A)}] (A) [left=-.2cm of B] {\includegraphics[height=5cm]{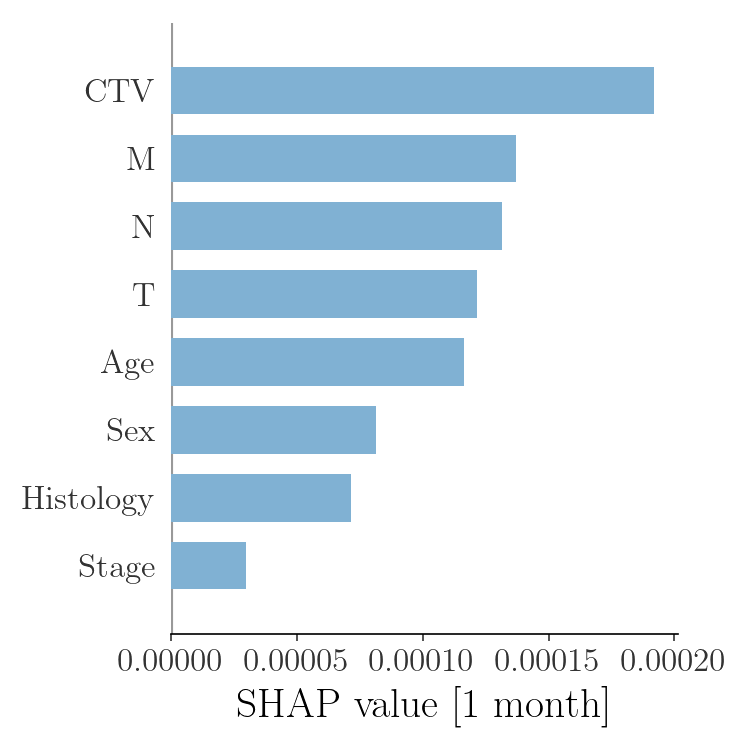}};
    \end{tikzpicture}}
    \caption{SHAP summary plots of features' contributions in the 3 models implemented with the 3 time units: Panel A) 1 month; Panel B) 1 year; Panel C) 2 years. The plots show the global feature importance by averaging the absolute SHAP values obtained for each patient and time represented in the output vector.}
    \label{fig:shap_values}
\end{figure}

Additionally, we performed an explainability analysis to better understand which features most influence the model's decisions. We used the SHAP method~\cite{bib:shap_values}, an explainability technique based on Shapley values, to estimate the contribution of each feature to the final output.
\figurename~\ref{fig:shap_values} shows the SHAP summary plots for the three models implemented with the 3 time units, i.e., 1 month, 1 year, and 2 years. In these plots, the order of the features is based on their importance in distinguishing the uncensored class. 
The higher the position of a feature in the plot, the more important it is.
To represent the global feature importance, we averaged the absolute SHAP values obtained for each patient and time represented in the output vector. 
Moreover, to get a general overview of the importance of the categorical features, we averaged the contributions of each category.
As we can see, in all time granularity models, the most relevant features for the task at hand refer to disease-related values such as the volume of the \textit{CTV} and the \textit{T}, \textit{N} and \textit{M} stages, which are measures of the severity of the tumour~\cite{bib:med}. 

\begin{figure*}[htpb]
    \centering
    \includegraphics[width=.8\textwidth, keepaspectratio]{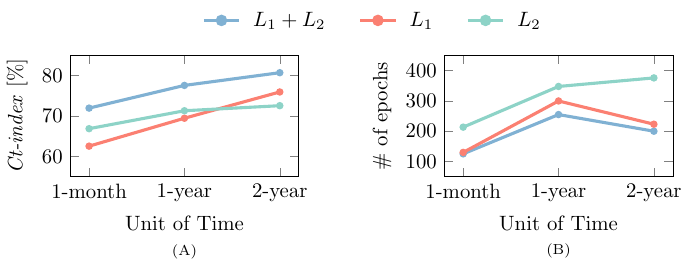}
    \caption{Ablation study of the two terms of the loss function proposed in \cite{bib:deephit}: (A)~Average performance (\textit{Ct-index}) and (B)~mean number of epochs to achieve convergence.}
    \label{fig:ablation}
\end{figure*}

Furthermore, we conducted an ablation study to gain a better understanding of the individual contributions of its two terms, $L_1$ and $L_2$. 
We examined whether both terms played a role in the model's final performance (\figurename~\ref{fig:ablation}.A) and in achieving convergence (\figurename~\ref{fig:ablation}.B).  
The figures depict the \textit{Ct-index} and the number of epochs, respectively, which clearly demonstrate that utilizing both terms together positively impacted the model's performance and the convergence time.

Finally, considering that the model is designed for a medical application, we analyzed its computational burden and inference speed. We evaluated the former in terms of the model's weight count, approximately $2.7$ million, which implies significantly longer training times compared to ML competitors, despite not requiring imputation.
However, since training would occur prior to deployment, the critical factor for practical application is the inference time, approximately $5\times10^{-3}$ seconds per sample, making it feasible for clinical practice.

\section{Conclusion}\label{sec:conclusion}

In this manuscript, we proposed a novel approach to address the problem of missing data in the context of survival analysis of NSCLC.
This task is usually tackled, in the literature, either by discarding the incomplete data samples or employing some imputation strategy, which could bias the final prediction.
Conversely, our method is able to ignore those values, taking into account the available ones only.
All the experiments are performed on the in-house dataset, CLARO~\cite{bib:CLARO}, consisting of the clinical information of 297 patients suffering from NSCLC.

From a clinical perspective, the development of reliable and accurate prognosis prediction tools, to be used prior to treatment initiation, is a critical and unmet need in clinical practice.
The availability of such tools would allow clinicians to tailor treatment strategies to the anticipated response, intensifying or descaling therapy as necessary based on the patient's prognosis. 

Experimental results show that our model outperforms conventional OS methods at various time units with different imputation techniques, bringing an increase in performance when missing features are present.

The results described so far and the limitations of this work suggest future directions worthy of investigation.
These include conducting additional experiments of our method on other OS datasets, as we are aware that validating a model on one dataset only, even if it is representative of a large population since it comes from a large Italian metropolis, is somewhat limiting, but obtaining medical datasets, preferably multicentric, is not always a straightforward process, particularly the raw version of the data in which imputation or exclusion of missing samples and features is not performed.
Therefore, in future works, we plan to extend the range of datasets, artificially generating different missing value scenarios on which to test our approach. 
Furthermore, this work considers clinical features only, thus having a reduced view of the patient's status, which could be expanded by taking into account other sources of information and types of data. 
Hence, another future work could expand our approach to include various types of data beyond tabular data (e.g., imaging data).
For a more seamless integration in the clinical setting, it could also be interesting to conduct further development to attempt to reduce the training time, by reducing the number of trainable weights, so as to accommodate eventual updates over time as new patient data are acquired.
Moreover, it could be interesting to explore the generalization ability of the proposed approach beyond the time-to-event analysis, thus directing future works in testing it in other task domains such as classification and regression.
In the meanwhile, we opted to make publicly available our GitHub repository\footnote{\url{https://github.com/cosbidev/OSTransformer}} in a way to make it possible to test our approach on other datasets.

\section*{Author Contributions}
\textbf{Camillo Maria Caruso}:
Conceptualization,
Data curation,
Formal analysis,
Investigation,
Methodology,
Software,
Validation,
Visualization,
Writing – original draft,
Writing – review \& editing;
\textbf{Valerio Guarrasi}:
Conceptualization,
Formal analysis,
Investigation,
Methodology,
Project administration,
Software,
Supervision,
Validation,
Writing – original draft,
Writing – review \& editing;
\textbf{Sara Ramella}:
Data curation,
Funding acquisition,
Resources;
\textbf{Paolo Soda}:
Conceptualization,
Funding acquisition,
Methodology,
Project administration,
Resources,
Supervision,
Writing – review \& editing.

\section*{Acknowledgment}
Camillo Maria Caruso is a Ph.D. student enrolled in the National Ph.D. in Artificial Intelligence, XXXVII cycle, course on Health and life sciences, organized by Università Campus Bio-Medico di Roma.

This work was partially founded by: i) Università Campus Bio-Medico di Roma under the program ``University Strategic Projects 2018 call'' within the project ``a CoLlAborative multi-sources Radiopathomics approach for personalized Oncology in non-small cell lung cancer (CLARO)''; 
ii) Università Campus Bio-Medico di Roma under the program ``University Strategic Projects'' within the project ``AI-powered Digital Twin for next-generation lung cancEr cAre (IDEA)''; 
iii) from PRIN 2022 MUR 20228MZFAA-AIDA (CUP C53D23003620008); 
iv) from PNRR MUR project PE0000013-FAIR.

Resources are provided by the National Academic Infrastructure for Supercomputing in Sweden (NAISS) and the Swedish National Infrastructure for Computing (SNIC) at Alvis @ C3SE, partially funded by the Swedish Research Council through grant agreements no. 2022-06725 and no. 2018-05973.



\bibliographystyle{elsarticle-num} 
\bibliography{bibliography.bib}





\end{document}